# A variational autoencoder approach for choice set generation and implicit perception of alternatives in choice modeling

**Rui Yao, Shlomo Bekhor**

Department of Civil and Environmental Engineering,

Technion – Israel Institute of Technology, Haifa 32000, Israel

Email: andyyao@campus.technion.ac.il,

     sbekhor@technion.ac.il

**Abstract**

This paper derives the generalized extreme value (GEV) model with implicit availability/perception (IAP) of alternatives and proposes a variational autoencoder (VAE) approach for choice set generation and implicit perception of alternatives. Specifically, the cross-nested logit (CNL) model with IAP is derived as an example of IAP-GEV models. The VAE approach is adapted to model the choice set generation process, in which the likelihood of perceiving chosen alternatives in the choice set is maximized.

The VAE approach for route choice set generation is exemplified using a real dataset. IAP-CNL model estimated has the best performance in terms of goodness-of-fit and prediction performance, compared to multinomial logit models and conventional choice set generation methods.

# 1. Introduction

The purpose of choice modeling is to model choices made by individuals among a set of available alternatives to them. It involves two main steps: choice set generation and modeling the choice from a given choice set. The choice set generation step is essential, since misspecification of the choice set considered by the individuals can cause inconsistency in coefficient estimations (Swait and Ben-Akiva, 1986; Ben-Akiva and Boccara, 1995; Frejinger et al., 2009). However, choice set generation is a nontrivial task in many aspects. For example, when the size of the full choice set is huge, enumerating all feasible alternatives is impractical. In addition, it is impossible to observe the true choice set considered by an individual before making a choice (Ben-Akiva and Boccara, 1995).

*Choice set generation*

Choice set generation is composed of alternative generation and choice set formation. Alternative generation methods can be divided into deterministic approach and stochastic approach (Frejinger et al., 2009). Deterministic alternative generation methods require pre-defined generation rules, and always generate the same set of alternatives for a given observation. When the full choice set is small, the generation rule can be simply enumerating all alternatives. If the full choice set is huge, typically, only a subset of feasible alternatives is generated. For example, in route choice models, a subset of routes is found by using variants of shortest path methods (e.g., link penalty, De La Barra et al., 1993; branch and bound, Prato and Bekhor, 2006). These deterministic methods are in general computational efficient, but one shortcoming is that there is no guarantee to reproduce the observed route, and may cause misleading interpretation of the estimation results.

Alternatively, stochastic methods are applied when the full choice set is large. Using again route choice modeling as an example, Frejinger et al. (2009), and Flötteröd and Bierlaire (2013) proposed explicit random walk methods for generating routes in road networks, which sample routes by randomly choosing outgoing links at each node. This random walk method was extended for implicit route choice modeling, which could include all route alternatives without enumeration, by assuming that individuals make successive decisions at each outgoing link nodes (Fosgerau et al., 2013; Mai et al., 2015; Nassir et al., 2019). These stochastic methods could avoid inconsistency in coefficient estimates, by assuming that individuals consider the full choice set. However, this assumption can be behaviorally questionable (Frejinger et al., 2009). Recently, Yao and Bekhor (2020) combine labeling method (Ben-Akiva et al., 1984) with sampling method for implicitly generating route alternatives, in which the behavioral aspect of alternative generation is captured by route characteristic clusters. Although their approach may be behaviorally reasonable, it lacks a systematically defined modeling framework.

After generating a set of alternatives, the choice set is then formed by deterministically including all or part of the generated alternatives, or probabilistically determined by a choice set formation model, which assigns the probability of individual considering a choice set (Manski, 1977). Deterministic choice set formation approach has the risk of inconsistent coefficient estimates when the full choice set is not used (Frejinger et al., 2009). Sampling correction terms have been derived for correcting the utilities when important sampling of alternatives is used, and to have unbiased coefficient estimates (Frejinger et al., 2009; Guevara and Ben-Akiva, 2013a, b; Lai and Bierlaire, 2015).

The probabilistic choice set formation approach is more general in the sense that all the combinations of different alternatives (choice sets) are considered. For example, Swait and Ben-Akiva (1987) generalized the elimination-by-aspect by introducing random constraints in the choice set generation process; Ben-Akiva and Boccara (1995) considered the latent choice sets, in which the alternative availability is modeled as a binary latent variable. Cascetta and Papola (2001) and Cascetta et al. (2002) simplified the probabilistic choice set formation approach, by assuming each alternative is associated with an implicit degree of availability/perception (IAP) measure. The resulted choice set is fuzzy, in which the composition of alternatives is probabilistically determined by the IAP, and explicitly enumerating of all possible alternative combinations is avoided. Probabilistic methods are more general, but specifying the choice set formation model is difficult, and it is often simplified. For example, the implicit availability/perception of alternative is modeled as a simple binomial logit model (Cascetta et al., 2002). Therefore, there is a need for developing a more general probabilistic choice set formation modeling framework.

*Choice models and adaptation of machine learning methods*

There exists a wide range of literatures for various choice models. For example, in the context of route choice modeling, multinomial logit (MNL) based models, such as, Path-size Logit (Ben-Akiva and Bierlaire, 1999), maintain the simple MNL structure with additional correction term in the systematic utility function accounting for correlations between the alternatives. Generalized extreme value (GEV) based models can capture the similarities between the alternatives in the random error part (Prato, 2009), for example, the link-nested logit (Vovsha and Bekhor, 1998) considers each link as a nest in the cross-nested structure. Fosgerau et al. (2013) proposed an implicit route choice model without explicitly generating routes a priori, by adapting a link-based formulation.

Recently, several choice models have adapted machine learning methods, because of their robust prediction power and automated feature learning ability. For example, Wang et al. (2020a, 2021) developed deep neural network models with dedicated architectures for synergizing discrete choice models, and showed the combination of machine learning method and random utility theory can enhance the prediction and robustness of the estimated models. Sifringer et al. (2020) proposed to use neural networks for learning new deep representations from the explanatory variables, and show these new representations increase the model prediction accuracy. Yao and Bekhor (2020) applied clustering for alternative generation, and used random forest to extract importance features for model specification, and also reported improvement in model prediction. Wang et al. (2020b) further illustrated how to extract econometric information (e.g., elasticity, substitution patterns) from the deep neural networks. For a recent review of applying machine learning methods in choice modeling, we refer interested readers to Van Cranenburgh et al. (2021).

One type of machine learning methods, deep generative models (e.g. variational autoencoder), which infer the underlying sample generation process, have proven their abilities for producing high quality images, texts, and sounds. However, to the best of our knowledge, these deep generative models have not been adapted for choice modeling, or applied for choice set generation.

*Objectives*

In summary, although several choice set generation models were developed in the literature, there are still outstanding points. Firstly, alternative generation is a challenging task. Implicit methods, which consider the full choice set can be behaviorally questionable (Frejinger et al., 2009), and other (explicit) methods may require defining searching criteria a priori. Moreover, these methods may not be able to reproduce the chosen alternatives, while it is required for model estimations.

Secondly, choice set formation is a nontrivial task. When the full choice set is huge, enumerating all its alternatives can be difficult. Deterministic choice set formation methods with subset of alternatives could cause biased coefficient estimates. Probabilistic choice set formation models, namely implicit availability/perception (IAP) models, can consider all alternative combinations is more general. However, only MNL model was derived for the IAP models, and simple IAP term was considered.

This paper focuses on modeling the choice set generation process, and particularly deals with large choice sets. We aim to fill some of the gaps mentioned above, by adapting the more general variational autoencoder (VAE) approach for choice set generation, and including the implicit perception of alternatives in choice modeling. The developed approach is a novel combination of machine learning methods and traditional discrete choice models, and is expected to bring the following contributions:

1) The generalized extreme value (GEV) model with implicit availability/perception of alternatives models is derived, with cross-nest logit model as an example.
2) The variational autoencoder is adapted for choice modeling, in which the VAE model learns the latent representation of the chosen alternative, and generates new alternatives, by maximizing the likelihood of perceiving observed chosen alternatives.
3) An application of the proposed general VAE choice set generation approach is applied for the route choice modeling.
4) The proposed variation autoencoder can connect with other machine learning models, and can be jointly trained in a machine learning pipeline.

## 2. Methodology

### 2.1 Implicit availability/perception of alternatives

The implicit availability/perception of alternatives fall in the category of probabilistic choice set formation models (Manski, 1977), in which the probability of individual $n$ choosing alternative $i$ is expressed as following:

$$p_n(i) = \sum_{D_n \in H_n} p_n(i|D_n) p_n(D_n) \tag{1}$$

where, $C_n$ is the set of all feasible alternatives of individual $n$, and $H_n$ is the set of all non-empty subsets of $C_n$, $p_n(i|D_n)$ gives the probability of individual $n$ choosing alternative $i$ given the choice set is $D_n$, and $p_n(D_n)$ is the probability that individual $n$ considers choice set $D_n$ given $C_n$. Equation (1) implies a high degree of complexity, since the number of different alternative combinations, i.e., the number of choice sets $D_n$, is very large. Cascetta and Papola (2001) and Cascetta et al. (2002) proposed a simplified approach to deal with the complex probabilistic choice set formation models, by implicitly including an alternative availability/perception measure, random variable $\ln[BC_n(i)]$, in the utility function $U_{in}$ as follows:

$$U_{in} = V_{in} + \ln[BC_n(i)] + \varepsilon_{in} \tag{2}$$

where, $V_{in}$ is the systematic utility and $\varepsilon_{in}$ is the random error term. By considering implicit availability/perception of alternatives, the resulting choice set $D_n$ is fuzzy, and thus, each alternative $j$ in $C_n$ could be perceived in $D_n$ with implicit availability $BC_n(j)$. The IAP multinomial logit (MNL) model can then be obtained as follows:

$$p_n^{MNL}(i) = \frac{BC_n(i) \cdot \exp(V_{in})}{\sum_{j \in C_n} BC_n(j) \cdot \exp(V_{jn})} = \frac{BC_n(i) \cdot \exp(V_{in})}{\sum_{j \in D_n} BC_n(j) \cdot \exp(V_{jn})}$$
$$= \frac{\exp(V_{in} + \ln[BC_n(i)])}{\sum_{j \in D_n} \exp(V_{jn} + \ln[BC_n(j)])} \tag{3}$$

Note that, the alternatives not perceived by individual $n$ are not included in the choice set $D_n$, i.e., $BC_n(j) = 0, \forall j \notin D_n$, and IAP-MNL model (equation 3) is equivalent to the MNL model derived using equation (2).

McFadden (1978) showed that the MNL, the nested logit and cross-nested logit (CNL) belongs to a more general class of generalized extreme value (GEV) choice model. The GEV choice probability can be written in a logit form (Ben-Akiva and Lerman, 1985):

$$p_n^{GEV}(i) = \frac{\exp(V_{in} + \ln[G_{in}])}{\sum_{j \in C_n} \exp(V_{jn} + \ln[G_{jn}])} \tag{4}$$

where, $G_{in}$ is the partial derivative of a generation function $G$, that is specific to each member of the MEV family. We propose to include the $BC_n(j)$ in the GEV choice probability as follows:

$$p_n^{GEV}(i) = \frac{BC_n(i) \cdot \exp(V_{in} + \ln[G'_{in}])}{\sum_{j \in C_n} BC_n(j) \cdot \exp(V_{jn} + \ln[G'_{jn}])} \tag{5}$$

where, $G'_{in}$ is the partial derivative of the generation function that includes $BC_n(j)$. Equation (5) can be seen as a generalization of equation (4), in which the availability of an alternative can take intermediate values $BC_n(j) \in [0,1]$, instead of only being binary. Assuming that all alternatives in $C_n$ can be (definitely) perceived by the individuals, that is, $BC_n(j) = 1, \forall j \in C_n$, equation (5) collapses to equation (4).

The CNL model can be formulated as a member of the GEV family. In the following, we derive equation (5) for the CNL model as an example, from the generation function $G^{CNL}$ with $M$ nests and implicit availability $BC_n(j)$ as follows:

$$G^{CNL} = \sum_{m=1}^{M} \left( \sum_{j \in C_n} BC_n(j) \cdot \alpha_{jm} e^{\mu_m V_{jn}} \right)^{\frac{\mu}{\mu_m}} = \sum_{m=1}^{M} \left( \sum_{j \in D_n} BC_n(j) \cdot \alpha_{jm} e^{\mu_m V_{jn}} \right)^{\frac{\mu}{\mu_m}} \quad (6)$$

where, $\mu > 0$ is the scale parameter for the model, $\mu_m$ is the scale parameter for nest $m$, $\alpha_{im} \in [0,1]$ is the inclusion parameter that captures the degree of membership of alternative $i$ belonging to nest $m$. We also assume $BC_n(j) = 0, \forall j \notin D_n$, and the properties of GEV generation functions are verified for equation (6) in the Appendix.

Then, the partial derivation of $G^{CNL}$, with respect to $\exp(V_{in})$ is:

$$\frac{\partial G^{CNL}}{\partial \exp(V_{in})} = \sum_{m=1}^{M} \left( BC_n(i) \cdot \mu \alpha_{im} e^{V_{in}(\mu_m - 1)} \left( \sum_{j \in D_n} BC_n(j) \cdot \alpha_{jm} e^{\mu_m V_{jn}} \right)^{\frac{\mu - \mu_m}{\mu_m}} \right)$$

$$= BC_n(i) \cdot \sum_{m=1}^{M} \left( \mu \alpha_{im} e^{V_{in}(\mu_m - 1)} \left( \sum_{j \in D_n} \alpha_{jm} e^{\mu_m V_{jn} + \ln[BC_n(j)]} \right)^{\frac{\mu - \mu_m}{\mu_m}} \right)$$

$$= BC_n(i) \cdot \mu {G'}_{in}^{CNL} \quad (7)$$

Note that, equation (7) is similar to the CNL sampling correction proposed by Guevara and Ben-Akiva (2013a), in which the $BC_n(j)$ is replaced by an expansion factor that compensates for non-sampled alternatives of the nests. Although the formulations of ${G'}_{in}^{CNL}$ are similar, the underlying assumptions are different. The sampling corrections assume the choice set $D_n$ is sampled by the modelers from $C_n$, while the IAP approach assumes the choice set is formed by individuals with each alternative associated with implicit availability/perception.

The choice probability of the CNL model with implicit availability/perception is:

$$p_n^{CNL}(i) = \frac{e^{V_{in}} \cdot G_{in}^{CNL}}{\mu \cdot G^{CNL}}$$

$$= \frac{BC_n(i) \cdot \exp(V_{in} + \ln[{G'}_{in}^{CNL}])}{\sum_{j \in D_n} BC_n(j) \cdot \exp(V_{jn} + \ln[{G'}_{jn}^{CNL}])} \quad (8)$$

Although we only explicitly derive for the IAP-CNL model, the proposed equation (5) is general, and the same derivation can be adapted to other members of the GEV family. We distinguish our IAP-CNL derivation from the IAP model proposed by Cascetta and Papola (2001) and Cascetta et al. (2002). Their approach considers $\ln[BC_n(j)]$ as part of the systematic utility function, and consequently $\ln[BC_n(j)]$ will be scaled by $\mu_m$ for each nest $m$. Instead, our proposed IAP-CNL model (equations 7) considers the implicit availability/perception $\ln[BC_n(j)]$ independent of nests, thus not scaled by $\mu_m$.

In contrast to Cascetta et al. (2002) that used pre-defined rules to generate the choice set, we propose a more general choice set generating approach that maximizes the likelihood of perceiving the chosen alternative into the choice set without defining the rules a priori, and jointly computes the implicit availability/perception $BC_n(j)$ using a more flexible approach, as described in the following subsection.

## 2.2 A variational autoencoder approach for choice set generation and implicit perception of alternatives

In this subsection, we propose a new approach based on variational autoencoders (VAE) for choice set generation and implicit perception of alternatives. The goal of the proposed approach is not only to perceive (or make available) the observed chosen alternative into the choice set $D_n$, but also to provide a more general modeling framework for implicit availability/perception $BC_n(j)$. Variational autoencoder (Kingma and Welling, 2014) provides a comprehensive and flexible framework for estimating the probability densities of the observations and generating new samples. To this end, the VAE is adapted in our paper for choice set generation and implicit perception of alternatives.

The autoencoder is composed of two parts: encoder and decoder. The encoder $\Phi$ is an inference model that maps the observed chosen alternative to a more compact lower-dimension latent representation. The decoder $\Theta$ is a generative model that produces new samples, given the latent representations. The purpose of an autoencoder model is to find the encoder and decoder such that the observation can be reproduced:

$$\Theta(\Phi(j)) \approx j \qquad (9)$$

By combining with variational Bayesian methods, the variational autoencoder model seeks to find two probability distributions for the inference process $\Phi$, and generative process $\Theta$, respectively, and use them to approximate the probability distribution for the underlying generation process. We adapt this general VAE method for choice set generation and implicit perception of alternatives.

An initial attempt to generate alternatives using aggregated information on diverse chosen alternatives was performed in Yao and Bekhor (2020) for route choice modeling. By aggregating information on the chosen routes, we could infer the characteristics of these perceived routes, and use this information to generate new alternatives. Although it was not applied systematically, the described generation process resembles the encoder-decoder method. Therefore, this motivated us to apply VAE models for choice set generation.

Hypothetically, if there exists a true distribution $q^*(j)$, which provides the likelihood of individuals perceiving alternative $j$ in the choice set, this distribution is equivalent to $BC_n(j)$. However, the true distribution $q^*(j)$ is not known, and only some samples from this true distribution can be obtained. For example, in the route choice case, we typically observe only the chosen alternatives. We are interested in approximating this true distribution using the observed samples. This problem then becomes a maximum (log)likelihood estimation problem for a parametric family of distributions $q_\theta(j)$ with parameters $\theta$. This aligns with our objective to maximize the likelihood of perceiving the chosen alternative into the choice set.

However, it is difficult to express $q_\theta(j)$, since the approximated distribution may not be analytically tractable. To cope with the potentially very complex distribution, we introduce some latent variables $z$, so that the joint distribution $q_\theta(j, z)$ can be defined as a product of simpler distributions. One possible interpretation for these latent variables is that the underlying alternative perception process is associated with some random constraints $z$ (Swait and Ben-Akiva, 1987; Ben-Akiva and Boccara, 1995).

The proposed VAE approach tries to reproduce the observations and generates new alternatives, by maximizing the likelihood of the chosen perceived alternatives. Following Kingma and Welling (2014), the log-likelihood function is derived as follows:

$$\log q_\theta(j) = \log \int \frac{p(z|j)}{p(z|j)} p(z) q_\theta(j|z) dz = \log \mathbb{E}_{p(z|j)} \left[ \frac{p(z) q_\theta(j|z)}{p(z|j)} \right] \tag{10}$$

where, the latent variable $z$ is described by the prior distribution $p(z)$. The likelihood of perceiving alternative $j$ from $C_n$ conditional on $z$ is $q_\theta(j|z)$, and the likelihood of inferring the latent variable $z$ given the chosen perceived route $j$ is the posterior distribution $p(z|j)$.

In the context of VAE models, $z$ is the compact latent representation of the alternatives, $q_\theta(j|z)$ corresponds to the generative model (decoder $\Theta$), and $p(z|j)$ corresponds to the inference model (encoder $\Phi$). Note that, this posterior distribution $p(z|j)$ is coupled through Bayes theorem:

$$p(z|j) = \frac{p(z) q_\theta(j|z)}{q_\theta(j)} \tag{11}$$

Equation (11) above cannot be evaluated analytically, except for very simple cases. To deal with this, the posterior distribution $p(z|j)$ can be approximated with a parametric family of distributions $p_\phi(z|j)$.

In Cascetta and Papola (2001) and Cascetta et al. (2002), $\ln[BC_n(j)]$ (or equivalently $\log q_\theta(j)$ in our case) is approximated by its average value $\mathbb{E}[\ln BC_n(j)]$ over all the individuals. In the context of VAE, this approximation can be interpreted as learning the alternative perception process from all the observations. Similarly, Monte Carlo simulation is applied for estimating $\log q_{BC}^\theta(j)$, and the estimator can be derived using Jensen's Inequality as follows (Burda et al., 2015):

$$\log q_\theta(j) = \log \mathbb{E}_{p_\phi(z|j)} \left[ \frac{1}{S} \sum_{i=1}^{S} \frac{p(z) q_\theta(j|z)}{p_\phi(z|j)} \right] \geq \mathbb{E}_{p_\phi(z|j)} \left[ \log \frac{1}{S} \sum_{i=1}^{S} \frac{p(z) q_\theta(j|z)}{p_\phi(z|j)} \right] = \mathcal{L} \tag{12}$$

Where $\mathcal{L}$ is the lower bound of $\log q_\theta(j)$. Then, by maximizing the lower bound $\mathcal{L}$, we are expecting to maximize $\log q_\theta(j)$ as well (given the bound is tight enough). For detailed properties and gradient derivation on this estimator, we refer interested readers to Burda et al. (2015).

Given the alternative $j$, the estimated $q_\theta(j|z)$ and $p_\phi(z|j)$, the log of implicit perception $BC_n(j)$ can be estimated using equation (13):

$$\ln[BC_n(j)] \approx \ln \frac{1}{S} \sum_{i=1}^{S} \frac{p(z)q_\theta(j|z)}{p_\phi(z|j)} \qquad (13)$$

Note that, the hyperparameter $S$ defines the number of random draws in the Monte Carlo for estimating the implicit perception of alternatives for a single alternative $j$, and $\mathcal{L}$ approaches $\log BC_n(j)$ as $S$ goes to infinite (Burda et al., 2015).

The following steps summarize the proposed choice set generation process:

- *Alternative generation steps:*
  1. Draw $z$ at random from the prior distribution $p(z)$
  2. Draw new alternative $j$ from the decoder $\Theta$, $q_\theta(j|z)$, given $z$ from step 1
- *Implicit perception estimation steps:*
  1. Draw $z$ from the encoder $\Phi$, $p_\phi(z|j)$, for the generated/chosen alternative $j$
  2. Draw from the decoder $\Theta$, $q_\theta(j|z)$, for the given $z$ from step 1
  3. Draw from prior distribution $p(z)$
  4. Compute $\widehat{BC}(j) = \frac{p(z)q_\theta(j|z)}{p_\phi(z|j)}$
  5. Repeat step 1-4 for $S$ times random draws
  6. Calculate $\ln BC_n(j)$ using equation (13)

In the following subsection, we provide an example of applying the proposed variational autoencoder for choice set generation and obtaining implicit perception in the context of route choice modeling.

**2.3 Application of the VAE approach for route choice modeling**

In this paper, we extend the data-driven route choice set generation approach proposed by Yao and Bekhor (2020), by considering implicit availability/perception of alternatives, and applying CNL model for capturing the similarities between alternatives. The data-driven approach infers route characteristics using the chosen routes, and distinguishes routes by their characteristic attributes. Therefore, alternative generation means sampling their attributes from the inferred route characteristic clusters.

As discussed in the previous subsection, the probability distributions $p_\phi(z|j)$ and $q_\theta(j|z)$ that models the underlying alternative perception process can be very complex. Kingma and Welling (2014) and Burda et al. (2015) suggest using neural networks for parameterizing these distributions, and showed the flexibilities and comprehensiveness of neural network models. Therefore, we adapt the neural network VAE approach for route choice set generation and alternative perception estimation, and estimate route choice models with implicit availability/perception of alternatives.

The application of the proposed VAE method for route choice modeling is shown in Figure 1.

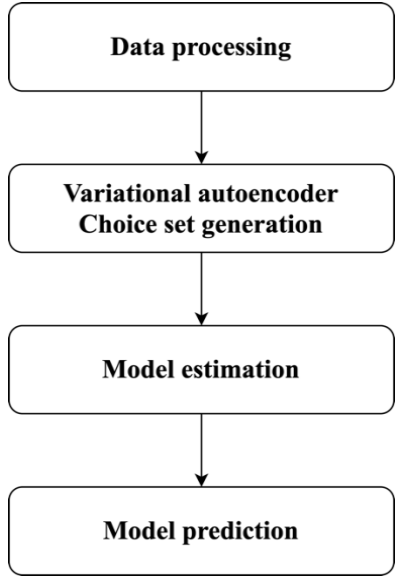

Figure 1 Procedures for applying the proposed VAE method for route choice modeling

The route characteristic attributes are in general non-negative, and the probability distribution $q_\theta(j|z)$ is often assumed to be log-normal, Gamma or truncated Normal in the literature (Nielsen and Frederiksen, 2006; Prato, 2009). Similarly, we assume $q_\theta(j|z)$ follows truncated Normal distribution as follows:

$$q_\theta(j|z) = \text{TruncatedNormal}(\mu_\theta(z), \sigma^2 I, 0, \infty) \tag{14}$$

where, the mean $\mu_\theta(z)$ of the truncated normal distribution is the decoder neural network, and $\theta$ are the weights of the neural network, variance $\sigma$ is a fixed hyperparameter, and $q_\theta(j|z)$ is bound from below at 0. And the posterior distributions $p_\phi(z|j)$ is assumed to follow a parametric normal distribution:

$$p_\phi(z|j) = N\left(\mu_\phi(j), \text{diag } \sigma_\phi(j)\right) \tag{15}$$

where, the mean $\mu_\phi(j)$ and variance $\sigma_\phi(j)$ of the normal distribution are the encoder neural networks, and $\phi$ are the weights of the neural network. Based on this assumption, the latent variable $z$ can be interpreted as cluster scores, capturing the membership of the alternative belonging to which cluster (nest); or as latent factors for the random constraints (Swait and Ben-Akiva, 1987; Ben-Akiva and Boccara, 1995). For simplicity, we assume that the prior distribution follows a standard normal distribution:

$$p(z) = N(\mathbf{0}, I) \tag{16}$$

The proposed neural network VAE model is shown in Figure 2. Herein, $j = \{x_1, \ldots, x_a\}$ are the characteristics attributes of an alternative $j$, $z = \{z_1, \ldots, z_b\}$ are the attributes of the latent variable.

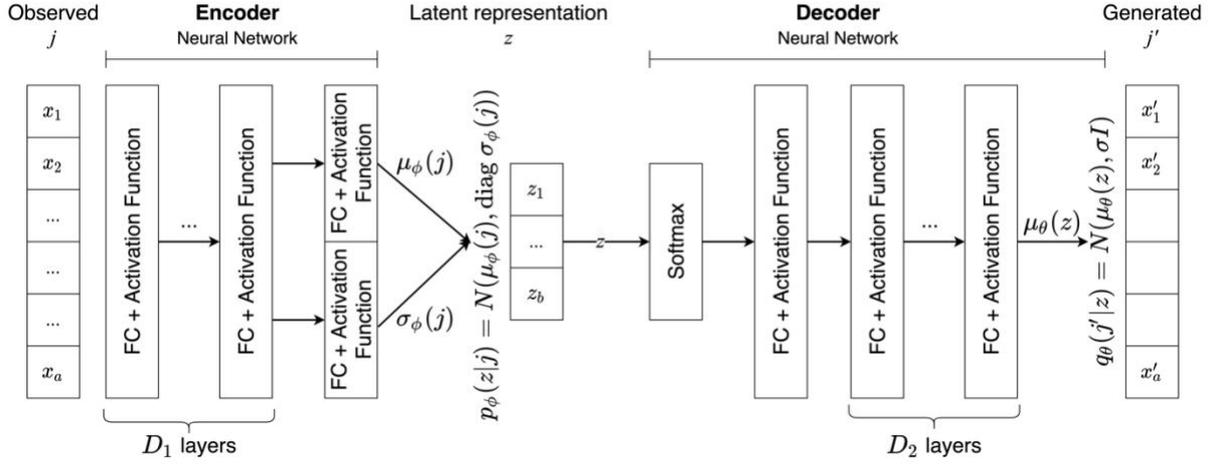

Figure 2 VAE neural network for route choice modeling

In the training phase of this neural network, the alternative attributes $j$ undergo transformation in the encoder neural network first, which then estimate the mean $\mu_\phi(j)$ and variance $\sigma_\phi(j)$ of the posterior distribution $p_\phi(z|j)$. Next, the latent attributes $z$ are sampled from this posterior distribution and passed to the decoder neural network.

On receiving the latent attributes, the decoder network first normalizes these attributes using the Softmax (MNL) function, and then use these attributes to estimate the mean $\mu_\theta(z)$ of the implicit perception of alternatives $q_\theta(j|z)$. Finally, a new sample is draw from $q_\theta(j|z)$. The VAE model parameters are estimated by maximizing the log-likelihood function in equation (12).

When the training phase is finished, we can generate new alternatives by drawing $z$ from the prior distribution $p(z)$, and pass it to the decoder network. The implicit perception of alternatives, $\ln BC_n(j)$, can then be easily obtained by passing the alternative (chosen or generated) to the VAE neural network and computing it using equation (13). Obtaining the implicit perception of alternatives is similar to training the VAE model, except that the parameters in the neural networks are fixed.

The fully connected (FC) layers considered in the proposed VAE model can exploit the automated feature learning capability of neural network and can be used as a universal approximator (Hornik, 1991; Wang et al., 2020). As discussed above, the latent variables $z$ can be interpreted as cluster scores. In this case, the Softmax function acts like a classifier for specifying the degree of membership $\alpha_{jm}$ of alternative $j$ to nest (cluster) $m$. The decoder neural network generates new samples given this membership information.

Given the generated choice set $D_n$, the implicit perception of alternatives $BC_n(j)$, and the degree of membership $\alpha_{jm}$ obtained using the above VAE model, we specify the systematic part of the utility function as follows:

$$V_{jn} = \sum_a \beta_a \cdot x_{a,jn} \qquad (17)$$

where, $\beta_a$ is the generic coefficient for attribute $a$, and $x_{a,jn}$ is the attribute value for alternative $j$ of individual $n$. By substituting equation (17) into equation (3) and equation (8) respectively, we can estimate route choice models with implicit availability/perception of alternatives for MNL and CNL using a training subset. The model prediction performance is evaluated using the testing subset with the estimated models.

## 3. Results

In this section, we present results of the proposed VAE approach for route choice modeling application described in subsection 2.3. The following subsections correspond to the procedure described in Figure 1.

### 3.1 Data processing

The dataset for the analysis is based on the Tel Aviv household travel survey data and map matched GPS trajectories. The survey collected information from 28,530 individuals and 265,815 trips over a 2-day period, with their GPS observation data using a designated mobile phone application. A detailed description of the respondent recruitment and data collection process can be found in Nahmias-Biran et al. (2018). After cleaning and filtering the GPS data, 5,002 car trips are map matched to a detailed planning network of Tel Aviv metropolitan area, which contains 8,583 nodes and 21,151 directed links. Main statistics of selected attributes are summarized in Table 1.

These selected route characteristic attributes are used in the neural network VAE model. In order to adequately apply these attributes for the neural network VAE model, there is a need to normalize them. While for model estimation, these normalized route characteristic attributes are converted to absolute values corresponding to each observation. For details on data preparation and conversion, we refer interested readers to Yao and Bekhor (2020).

Table 1 Main statistics of selected route characteristic attributes

| Attributes | Mean | Std. |
|---|---|---|
| Route average intersection time (over all intersections) | 0.18 | 0.07 |
| Route length detour (ratio to shortest path length) | 1.11 | 0.21 |
| Route time detour (ratio to fastest path time) | 1.08 | 0.17 |
| Route average number of links (per km) | 4.42 | 2.21 |
| Route city node percentage (ratio of num of intersections in the city center to all intersections) | 0.1 | 0.19 |
| Route percentage delay (ratio of delay to free-flow travel time) | 1.85 | 0.61 |
| Route highway/expressway percentage (of total distance) | 0.71 | 0.29 |
| Route left turn percentage (of total number of intersections) | 0.11 | 0.09 |
| Route average operating cost (per km) | 1.06 | 0.39 |

Furthermore, the 5,002 observations are randomly split into a training subset of size 4,000 and another test subset with the remaining 1,002 observations. In order to have a fair comparison between different models, the same training subset and test subset will be used in different models.

## 3.2 Hyperparameters for VAE choice set generation

The flexibility of neural network models is associated with the hyperparameters defining the actual architecture of the neural networks. These hyperparameters could affect the performance, in our case the implicit perception estimation, of the neural network VAE model. We list the hyperparameters and their value ranges in Table 2.

Table 2 Hyperparameters of the neural network VAE model

| Hyperparameters | Values |
| --- | --- |
| *Fixed hyperparameters* | |
| Initialization | He initialization |
| Activation function *for the hidden layers* | Tanh |
| Decoder variance $\sigma$ | 1.0 |
| | |
| *Varying hyperparameters* | |
| Latent space dimension $|z|$ | [1, 2, **3**, 4, 5, 6, 7, 8, 9, 10] |
| Encoder number of hidden layers $D_1$ | [0, 1, **2**, 3, 4, 5, 6] |
| Decoder number of hidden layers $D_2$ | [0, 1, 2, **3**, 4, 5, 6] |
| Batch normalization | [**True**, False] |
| Mini-batch size | [50, **100**, 200, 500, 1000] |
| Learning rate | [0, 0.1, 0.01, **$10^{-3}$**, $10^{-5}$] |
| Number of random draws $S$ in Monte Carlo | [50, **100**, 200, 500, 1000] |
| Number of iterations | [500, 1000, 5000, **10000**, 20000] |

The hyperparameters are defined in two types, the fixed hyperparameters and varying parameters. For the fixed hyperparameters, we apply the commonly used He initialization (He et al., 2015) method for initializing the parameters of the neural networks in VAE. The choice of Activation function as hyperbolic tangent function (Tanh), and the Decoder variance $\sigma$ as 1.0 is related to the data processing step, in which the attributes are standardized for aggregating information gathered from routes of different OD pairs (Yao and Bekhor, 2020).

The other hyperparameters can be specified with different values, and could result in different model performance, we briefly introduce these varying hyperparameters in the following. In the context of the route choice modeling application, Latent space dimension $|z|$ defines the number of route nests (clusters); Number of hidden layers $D_1$ and $D_2$ controls the complexity of the encoder and decoder neural networks; Batch normalization is related to normalization of each batch in the stochastic gradient descent (SGD) optimization; Mini-batch size controls the number of samples used for SGD in one optimization step; Learning rate determines the step size at each SGD optimization step; $S$ defines the number of random draws to approximate the implicit perception in Monte Carlo simulation; and Number of iterations defines the maximum number of iterations for training the neural networks.

For these varying hyperparameters, we would like to identify the neural network VAE configuration (combination of hyperparameters in Table 1) with a high implicit availability/perception for the chosen alternatives, i.e., $\sum_n \ln BC_n(i)$. We apply the random hyperparameter searching method (Bergstra and Bengio, 2012) for selecting such configuration. In this paper, 50 sets of neural network VAE configurations are used for training, we summarize the log-likelihood of the test set ($\sum_n \ln BC_n(i)$) for the top 30 models in Figure 3.

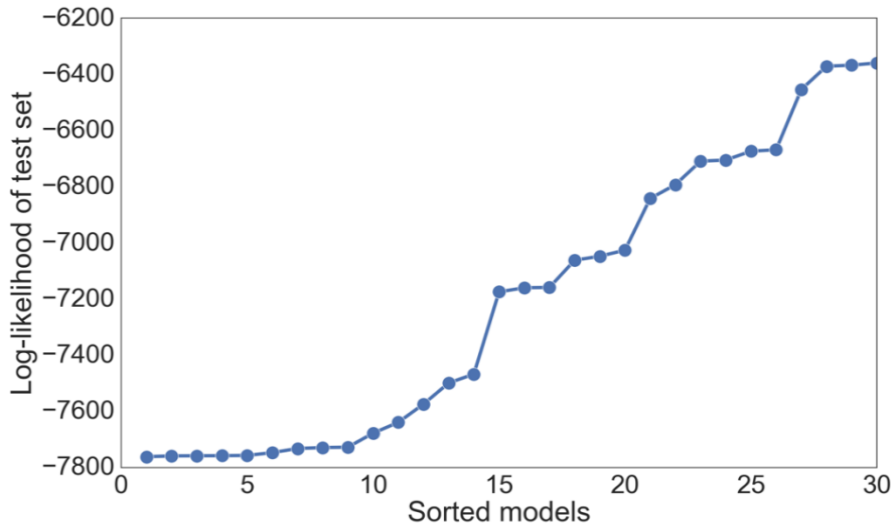

Figure 3 Hyperparameter searching results

In order to have better generalizability, and generate new alternatives with higher implicit perception, we select the model with the highest log-likelihood of the test set as our final model for route choice set generation and alternative perceptions (IAP) estimation. The selected hyperparameter configuration is marked in bold in Table 2. In the following subsection, we generate alternatives and estimate route choice models with this selected neural network VAE model.

**3.3 Route choice model estimation and prediction**

In this subsection, we present estimation results of the route choice models using the training subset, and prediction performance of the estimated model using the testing subset. All the models presented here are specified using equation (17), with explanatory variables listed in Table 1 (converted from normalized attributes to absolute attributes).

The route choice sets are generated using neural network VAE model with the selected hyperparameter configuration, and the implicit alternative perceptions $\ln BC_n(j)$ are estimated. The choice set size is assumed to have 20 alternatives (as suggested in Bekhor et al., 2006). Four types of route choice models with VAE choice set generation are estimated: 1) MNL; 2) CNL; 3) MNL with implicit alternative perceptions; and 4) CNL with implicit alternative perceptions. In addition, 3 replications of the generation–estimation procedure is performed. The estimation results are summarized in Table 3, in which values in brackets show the t-test values against 0. Results indicate that most coefficients have significant explanatory power, in terms of the t-test values. The additional IAP terms provide coefficient estimates close to the ones without IAP (as we do not expect dramatic changes in the estimates by introducing the IAP term). Moreover, the neural network VAE model produces consistent coefficient estimates across different replication runs for each type of models.

Table 3 Estimation results of route choice models with VAE choice set generation

| Attributes | MNL | | | CNL | | | MNL with IAP | | | CNL with IAP | | |
| --- | --- | --- | --- | --- | --- | --- | --- | --- | --- | --- | --- | --- |
| | Run | | | Run | | | Run | | | Run | | |
| | 1 | 2 | 3 | 1 | 2 | 3 | 1 | 2 | 3 | 1 | 2 | 3 |
| Route intersection time | -0.21 (-17.20) | -0.22 (-18.50) | -0.20 (-18.00) | -0.30 (-19.60) | -0.30 (-18.90) | -0.27 (-17.70) | -0.15 (-17.00) | -0.15 (-18.30) | -0.14 (-17.80) | -0.20 (-19.00) | -0.20 (-18.70) | -0.18 (-17.30) |
| Route length | -0.07 (-8.57) | -0.07 (-8.65) | -0.07 (-8.99) | -0.03 (-3.45) | -0.06 (-6.23) | -0.05 (-5.50) | -0.05 (-8.57) | -0.05 (-8.47) | -0.05 (-8.88) | -0.03 (-4.12) | -0.04 (-6.05) | -0.03 (-5.38) |
| Route time | -0.95 (-40.70) | -0.94 (-40.40) | -0.95 (-40.80) | -1.10 (-38.20) | -1.11 (-38.30) | -1.08 (-38.00) | -0.63 (-37.80) | -0.65 (-40.30) | -0.66 (40.80) | -0.72 (-36.20) | -0.75 (-37.80) | -0.73 (-37.70) |
| Route number of links | 0.13 (8.04) | 0.16 (9.77) | 0.14 (8.88) | 0.26 (12.50) | 0.30 (14.10) | 0.28 (13.50) | 0.08 (7.18) | 0.11 (9.90) | 0.10 (9.12) | 0.17 (11.50) | 0.20 (14.00) | 0.19 (13.40) |
| Route city node num | 0.10 (22.00) | 0.09 (19.90) | 0.08 (19.10) | 0.12 (21.80) | 0.10 (20.40) | 0.10 (19.50) | 0.07 (20.30) | 0.06 (19.90) | 0.06 (18.90) | 0.08 (21.00) | 0.07 (20.10) | 0.07 (18.50) |
| Route delay | 0.01 (2.09) | 0.02 (3.07) | 0.02 (4.18) | 0.00 (0.58) | 0.01 (1.88) | 0.03 (3.33) | 0.01 (1.72) | 0.01 (2.96) | 0.02 (4.12) | 0.00 (0.21) | 0.01 (1.80) | 0.02 (3.26) |
| Route highway/expressway length | 1.17 (59.20) | 1.16 (59.50) | 1.15 (58.90) | 1.21 (52.60) | 1.22 (52.80) | 1.20 (51.90) | 0.78 (55.80) | 0.80 (59.70) | 0.80 (58.90) | 0.81 (50.40) | 0.83 (52.80) | 0.82 (51.80) |
| Route left turn num | 0.86 (46.60) | 0.82 (45.90) | 0.84 (45.50) | 0.84 (40.20) | 0.79 (38.10) | 0.78 (36.40) | 0.56 (43.10) | 0.56 (46.50) | 0.58 (45.60) | 0.55 (37.60) | 0.54 (38.60) | 0.53 (36.40) |
| Route operating cost | -0.02 (-1.60) | 0.00 (-0.28) | -0.01 (-0.91) | 0.10 (8.12) | 0.13 (9.74) | 0.09 (7.35) | -0.01 (-1.63) | 0.00 (-0.38) | -0.01 (-0.80) | 0.06 (7.57) | 0.08 (9.33) | 0.06 (7.18) |

We compare the performance of our proposed neural network VAE model for choice set generation, against two conventional models with choice sets constructed using link penalty method. These two additional models are specified using equation (17) as well, and assumed to follow MNL and CNL respectively, in which the CNL model is specified with link-nested structure (Vovsha and Bekhor, 1998). For details on the choice set generated using link penalty, we refer to Yao and Bekhor (2020b), and their estimation results are shown in the Appendix

Table 7.

The model prediction performances, in terms of log-likelihood, are evaluated using the testing subset with the estimated models. The Goodness-of-fit and prediction performance results are shown in Table 4. Remind that, the training subset has 4,000 observations, and the testing subset has 1,002 observations.

Table 4 Goodness-of-fit and prediction performance results

| Choice set generation method | | | **Link penalty** | | **Neural network VAE approach** | | | |
|---|---|---|---|---|---|---|---|---|
| | | Log-likelihood | MNL | CNL | MNL | CNL | MNL with IAP | CNL with IAP |
| Number of coefficients | | | 9 | 9 | 9 | 9 | 9 | 9 |
| Run 1 | Training subset | $LL(\beta=0)$ | -10010.79 | -9947.46 | -11982.93 | -10404.19 | -9646.49 | -8550.14 |
| | | $LL(\hat{\beta})$ | -5319.28 | -5304.10 | -3578.16 | -3344.21 | -3417.99 | -3144.83 |
| | Testing subset | $LL(\hat{\beta})$ | -1011.96 | -992.24 | -972.16 | -904.41 | -933.14 | -858.84 |
| Run 2 | Training subset | $LL(\beta=0)$ | - | | -11982.93 | -11982.93 | -12058.65 | -9613.64 |
| | | $LL(\hat{\beta})$ | | | -3611.42 | -3644.96 | -3646.35 | -3470.59 |
| | Testing subset | $LL(\hat{\beta})$ | | | -797.92 | -848.07 | -846.40 | -810.05 |
| Run 3 | Training subset | $LL(\beta=0)$ | - | | -11982.93 | -11982.93 | -12063.52 | -9587.77 |
| | | $LL(\hat{\beta})$ | | | -3644.96 | -3611.42 | -3609.04 | -3388.46 |
| | Testing subset | $LL(\hat{\beta})$ | | | -848.07 | -797.92 | -792.27 | -795.92 |

As shown in Table 4, the choice models estimated with choice set generated using neural network VAE model outperform models estimated with conventional link penalty choice set generation method, in terms of final model log-likelihood for the training subset and testing subset. This means that the neural network VAE model could generate alternatives that are more likely to be perceived, which provide additional information for model estimation.

Moreover, the CNL models with implicit alternative perceptions outperform all other models in terms of model estimation and prediction. This is because the derived IAP-CNL model captures the similarities between alternatives, and the implicit availability/perception $BC_n(j)$ dis-utilizes/corrects the alternatives that are less likely to be perceived by individual $n$.

Note that, similar to Yao and Bekhor (2020), in the neural network VAE model, the normalized route characteristic attributes dataset is used, while in model estimation/prediction step the absolute route attribute dataset is used. The above procedure is expected to avoid endogeneity issues.

## 4. Discussion

*Realization of the choice set*

The proposed VAE approach generalizes the approach proposed by Cascetta and Papola (2001) and Cascetta et al. (2002), in which alternatives are associated with an implicit degree of availability/perception, and thus, the resulting choice set is fuzzy.

One way to interpret this fuzziness is from the individual's perspective. For example, individuals may not perceive an alternative even if it is actually feasible. This is typically the case when the feasible full choice set is very large. For example, individuals may only perceive a subset of all the feasible routes, and choose among the alternatives in the subset.

Another way to interpret the fuzziness is from the modeler's perspective. Since the "true" choice set considered by the individuals typically cannot be observed, the IAP also captures the modeler's belief of certain alternative being considered by the individuals. In this case, the alternatives generated by the VAE model can be seen as the expectance of the fuzzy choice set. Consequently, the "actual" (fuzzy) alternative considered by the individual can be realized from the full choice set with mean and variance obtained from the VAE model, for a given level of confidence.

In the context of route choice modeling, the realization of fuzzy alternatives means searching for a route with attributes within a confidence interval, for a given origin-destination pair. This search can be done using well-known methods, for example, k-shortest paths, link elimination, random walk, etc. An example of route alternative realization is illustrated in Appendix Figure 4.

*Consistency in coefficient estimates*

We are interested in examining the consistency in coefficient estimates of our proposed VAE approach. For this task, we performed a simulated observation experiment similar to Frejinger et al. (2009). Accordingly, we specify a model with only Route length, Route highway/expressway length, and Route city node num, whose true coefficients are assumed to be $\beta_{\text{Route length}} = -1.5$, $\beta_{\text{Route highway/expressway length}} = 1.5$, and $\beta_{\text{Route city node num}} = 0.5$.

For this exercise, one observation was randomly selected from our dataset, and used to simulate 1000 observations, each one with 20 alternative routes. The same choice set generation procedure described in 2.2 is applied for Experiment 2. In addition, two experiments with different IAP ranges are conducted to examinate the impact of $BC_n(j)$ on model estimation. Note that, alternatives exceeding the defined range are purposely discarded during choice set generation, and re-generated until the observation has 20 alternatives. The experiments are summarized as follows:

- Experiment 1: IAP of the alternatives are relatively lower (alternatives with $BC_n(j) > 0.001$ are purposely discarded)

- Experiment 2: the same choice set generation procedure used for estimating the route choice models in 0 (all the alternatives are randomly drawn from VAE and kept)
- Experiment 3: the average IAP of the alternatives are relatively higher (alternatives with $BC_n(j) < 0.001$ are purposely discarded)

The estimation results for the three experiments are shown in Table 5, in which the estimated coefficient $\hat{\beta}$, the standard error, t-test on the estimated coefficient against zero, and against the preset true coefficient values are included.

Table 5 Simulation experiment results

| Attributes | | Experiment 1 (Low IAP) | Experiment 2 (Random) | Experiment 3 (High IAP) |
|---|---|---|---|---|
| Route length $\beta = -1.5$ | $\hat{\beta}$ | -1.62 | -1.68 | -1.86 |
| | Std. | 0.381 | 0.151 | 0.158 |
| | t-test(0) | -4.26 | -11.10 | -11.80 |
| | t-test(-1.5) | -0.31 | -1.19 | -2.28 |
| Route highway/expressway length $\beta = 1.5$ | $\hat{\beta}$ | 1.68 | 1.57 | 1.73 |
| | Std. | 0.335 | 0.122 | 0.128 |
| | t-test(0) | 5.01 | 12.80 | 13.50 |
| | t-test(1.5) | 0.54 | 0.57 | 1.80 |
| Route city node num $\beta = 0.5$ | $\hat{\beta}$ | 0.46 | 0.57 | 0.60 |
| | Std. | 0.089 | 0.049 | 0.047 |
| | t-test(0) | 5.23 | 11.70 | 12.90 |
| | t-test(0.5) | -0.39 | 1.51 | 2.21 |

For all three experiments, all the coefficients estimated are significantly different from 0 at a 5% significance level (critical value: 1.96), while the coefficients estimated in experiment 1 are less significant to the other two experiments. The implicit availability/perception of the chosen alternative is significantly higher than the unchosen alternatives (as designed for experiment 1), which implies $BC_n(i)$ has stronger impact on the alternative utilities.

In terms of estimation consistency compared to the preset true values, the coefficient estimated in experiment 1 and 2 are not significantly different from the true values. While for experiment 3, the coefficients are biased from its true value at a 5% significance level. This could be caused by the implicit availability/perception of the chosen alternative is lower than the unchosen alternatives (as designed for experiment 3), and deteriorate coefficient estimates from their true values.

Among these three experiments, the model estimated in experiment 2, with all the alternatives randomly drawn from the VAE model, outperforms other models. Experiment 2 not only provides significant coefficient estimates (against 0), but also the coefficient estimates are not biased from its preset true values (at a 5% significant level). Results also confirm the importance of properly generating choice sets, as misspecification of the choice set can result in biased model estimations.

*Runtime performance*

Furthermore, we are also interested in the runtime performances of different models (Table 6). All the runtime performances of different models are obtained using a 6-core PC with GPU.

Table 6 Runtime performance

| VAE training time* [min] | Choice set generation time [min] | | Model estimation time [min] | | | | | |
|---|---|---|---|---|---|---|---|---|
| | | | Link penalty choice set | | VAE choice set | | | |
| | VAE* | Link penalty | MNL | CNL | MNL | CNL | MNL with IAP | CNL with IAP |
| 37.68 | 5.51 | 15.83 | 0.51 | 84.37 | 0.79 | 1.38 | 0.81 | 1.41 |

*: VAE model training and VAE choice set generation is performed on GPU.

Results show that, the VAE choice set generation time is lower than the conventional link penalty method. Note that, the VAE training is a one-time procedure, and can be applied for different choice set generation tasks. For model estimation times, all the MNL models have similar estimation time, while CNL model estimation times are higher than MNL models. The estimation time of the link-nested CNL model with link penalty choice set is significantly longer than CNL models with VAE choice set. This is because the VAE choice set has a more compact nesting structure, i.e., 3 route characteristic clusters, than the link-nested structure (625 link nests), and consequently the model estimation time is shorter.

## 5. Summary and Conclusions

In this paper, we derive the IAP-GEV model, and specifically, the IAP-CNL model as an example. A novel approach adapting variational autoencoders for choice set generation and IAP in choice modeling is proposed. The VAE approach combines variational Bayesian methods with autoencoder to approximate the probability distribution for the underlying choice set generation process, by maximizing the likelihood of perceiving the chosen alternatives in the choice set.

The VAE first infers the alternatives to a more compact lower-dimension latent representation using the encoder model, then generates new alternatives with the decoder model given the latent representations. The choice set is implicitly generated by sampling latent representations from the latent space, and passing to the decoder model. Moreover, the IAP measure can be obtained from the VAE model when generating new alternatives.

The proposed general VAE approach is applied to model route choice in this paper. Specifically, the probability distributions in the applied VAE models are parameterized using neural networks. The latent representations in the VAE model are interpreted as route characteristic clusters (nests), which captures the similarities between the routes. Several model structures are estimated: MNL, CNL, IAP-MNL, and IAP-CNL, and their prediction performances are also evaluated.

Estimation results show that the models with choice set generated using the VAE approach outperform models with conventional choice sets, both in terms of model goodness of fit and prediction performance. In particular, the IAP-CNL model that captures the correlation between alternatives and considers the degree of perceptions of alternatives in the choice set, has the best performance. Methods for realizing alternatives from the fuzzy choice set are also suggested. The consistency of the proposed approach is also verified using simulated observation experiments.

The methodology proposed in this paper is general and can be applied for other choice modeling problems, such as destination choice problems, or parking location choice problems. The proposed VAE approach can connect with other machine learning models, such as the deep neural network adapted for choice modeling (Wang et al., 2020). In the context of route choice models, link-based traffic assignment models incorporating VAE implicit availability/perception will be developed in future research.

# 7. Appendix

**Theorem 1.** The following conditions are sufficient for equation (6) to define a GEV generation function:

1. $\alpha_{jm} \geq 0, \forall j, m,$
2. $\sum_m \alpha_{jm} > 0, \forall j,$
3. $\mu > 0,$
4. $\mu_m > 0, \forall m,$
5. $\mu_m \geq \mu, \forall m,$
6. $BC_n(j) > 0, \forall j \in D_n$

**Proof:** Adapted from Bierlaire (2006), under these assumptions, equations (6) can be verified for the four properties of GEV generation functions.

1. $G^{CNL}$ is obviously non-negative
2. $G^{CNL}$ is homogeneous of degree $\mu$:

$$G^{CNL}(\beta \exp(V_{in})) = \sum_{m=1}^{M} \left( \sum_{j \in D_n} \beta^{\mu_m} BC_n(j) \cdot \alpha_{jm} e^{\mu_m V_{jn}} \right)^{\frac{\mu}{\mu_m}}$$

$$= \sum_{m=1}^{M} \beta^{\mu} \left( \sum_{j \in D_n} BC_n(j) \cdot \alpha_{jm} e^{\mu_m V_{jn}} \right)^{\frac{\mu}{\mu_m}} = \beta^{\mu} \cdot G^{CNL} \quad (18)$$

3. The limit properties hold from both assumptions 2, and 6, which guarantee that there is at least one non-zero coefficient $\alpha_{jm}$ for each alternative $j$ perceived by the individuals ($BC_n(j) > 0, \forall j \in D_n$).

$$\lim_{\exp(V_{in}) \to \infty} G^{CNL} = \lim_{\exp(V_{in}) \to \infty} \sum_{m=1}^{M} \left( \sum_{j \in D_n} BC_n(j) \cdot \alpha_{jm} e^{\mu_m V_{jn}} \right)^{\frac{\mu}{\mu_m}}$$

$$= \sum_{m=1}^{M} \left( \lim_{\exp(V_{in}) \to \infty} \left( \sum_{j \in D_n} BC_n(j) \cdot \alpha_{jm} e^{\mu_m V_{jn}} \right)^{\frac{\mu}{\mu_m}} \right) = \infty \quad (19)$$

4. The condition for the sign of the derivatives is obtained, by adapting the Lemma proved by Bierlaire (2006), in which the $k^{th}$ partial derivatives with respect to $k$ distinct alternatives with indices $\{i_1, \ldots, i_k\}$ is:

$$\frac{\partial G^{CNL}}{\partial \exp(V_{i_1 n}) \ldots \partial \exp(V_{i_k n})}$$

$$= \sum_{m=1}^{M} \left( \mu_m^k \prod_{l \in \{i_1, \ldots, i_k\}} (BC_n(l) \alpha_{lm} e^{V_{in}(\mu_m - 1)}) \prod_{l=0}^{k-1} \left( \frac{\mu}{\mu_m} - l \right) y_m^{\frac{\mu - k\mu_m}{\mu_m}} \right) \quad (20)$$

where,
$$y_m = \sum_{j \in D_n} BC_n(j) \cdot \alpha_{jm} e^{\mu_m V_{jn}} \quad (21)$$

Then, the signs of the derivatives are shown in three cases:

a) For $k = 1$: it is obvious equation (7), $\frac{\partial G^{CNL}}{\partial \exp(V_{in})} \geq 0$

b) For $k > 1$ and $\mu_m = \mu$:
$$\frac{\partial G^{CNL}}{\partial \exp(V_{i_1 n}) \ldots \partial \exp(V_{i_k n})} = 0 \quad (22)$$

since, for $l = 1$, $\left(\frac{\mu}{\mu_m} - l\right) = 0$, and thus $\prod_{l=0}^{k-1} \left(\frac{\mu}{\mu_m} - l\right) = 0$.

c) For $k > 1$ and $\mu_m > \mu$, if $l > 0$:
$$\frac{\mu}{\mu_m} - l < 0 \quad (23)$$

And the signs of the derivatives are determined by
$$\prod_{l=0}^{k-1} \left(\frac{\mu}{\mu_m} - l\right) \begin{cases} \geq 0, if\ k\ is\ odd \\ \leq 0, if\ k\ is\ even \end{cases} \quad (24)$$

Therefore,
$$\frac{\partial G^{CNL}}{\partial \exp(V_{i_1 n}) \ldots \partial \exp(V_{i_k n})} \begin{cases} \geq 0, if\ k\ is\ odd \\ \leq 0, if\ k\ is\ even \end{cases} \quad (25)$$

∎

1 Table 7 Estimation results of link penalty models

| Attributes | MNL | CNL |
| --- | --- | --- |
| Route intersection time | 0.15 (6.71) | 0.13 (7.25) |
| Route length | -0.31 (-12.00) | -0.24 (-11.60) |
| Route time | -0.92 (-19.10) | -0.74 (-19.20) |
| Route number of links | 0.14 (5.35) | 0.11 (4.96) |
| Route city node num | -0.07 (-7.46) | -0.05 (-7.55) |
| Route delay | -0.27 (-6.50) | -0.23 (-6.74) |
| Route highway/expressway length | 0.44 (18.40) | 0.35 (18.60) |
| Route left turn num | -0.27 (-21.90) | -0.22 (-21.60) |
| Route operating cost | -0.08 (-9.46) | -0.07 (-9.56) |

Figure 4 Route alternative realization (routes with length within the confidence interval is marked in green, and the realized route alternative is marked in red.)